# Automated 3D reconstruction of LoD2 and LoD1 models for all 10 million buildings of the Netherlands

Ravi Peters, Balázs Dukai, Stelios Vitalis, Jordi van Liempt, Jantien Stoter
3D Geoinformation group, Delft University of Technology, Faculty of Architecture and the Built Environment, Department of Urbanism, Julianalaan 134, Delft 2628BL – The Netherlands
{r.y.peters| b.dukai | s.vitalis | j.n.h.vanLiempt | j.e.stoter}@tudelft.nl

**ABSTRACT**
In this paper we present our workflow to automatically reconstruct 3D building models based on 2D building polygons and a LiDAR point cloud. The workflow generates models at different levels of detail (LoDs) to support data requirements of different applications from one consistent source. Specific attention has been paid to make the workflow robust to quickly run a new iteration in case of improvements in an algorithm or in case new input data become available. The quality of the reconstructed data highly depends on the quality of the input data and is monitored in several steps of the process. A 3D viewer has been developed to view and download the openly available 3D data at different LoDs in different formats. The workflow has been applied to all 10 million buildings of The Netherlands. The 3D service will be updated after new input data becomes available.

## 1. INTRODUCTION

3D city models are widely used in urban applications. The outcomes of such applications serve as input for planning and decision-making processes that aim at making cities cooler, sustainable, better accessible, greener, $CO_2$-neutral, etc (Biljecki et al., 2016; Deren et al., 2021). Models of buildings are prominent objects in these models. The building models can be generated at different Levels of Detail (LoD). Taking the terminology of CityGML, a building can be modelled at four main levels of detail for the outer shell of the building, i.e., LoD0, LoD1, LoD2, and LoD3, and at LoD4 for the interior of the building (OGC, 2012; Kutzner et al, 2020). Each of these four CityGML LoDs can be further refined (Biljecki et al., 2016; Sun et al 2019).A higher level of detail is often preferred over a lower one, since building models at higher LoDs look closer to reality. However, higher levels of detail are more complicated (and therefore more expensive) to acquire, because it is harder to reconstruct them in an automated manner from available source data. In addition, using models at higher levels of detail in spatial analysis does not automatically lead to better results (Biljecki et al, 2018) while at the same time too much detail may have a negative impact on performance. Therefore, for some applications it is better to avoid too much, irrelevant detail.The LoD of a 3D city model is therefore driven by the specific data requirements of the urban application for which it is built (see also Section 2.1). However, the highest achievable LoD is also restricted by the available source data and the employed reconstruction method (see also Section 2.2).While many 3D city models exist for various parts of the Netherlands, they are often generated for relatively small areas, are using different reconstruction methods and are based on different source data. Furthermore, the update cycles are different and the level of detail is also different because they are collected for different applications.
This can result in inconsistencies between 3D city models of the same area. There may be discrepancies between the geometries of building models like the geometry or height of the footprint. Also, the reference heights for the same building might differ over data sets since the heights may represent different references (e.g. gutter, ridge, maximum) or the reference heights are based on different statistical calculations. In addition, buildings (or building parts) available in one data set might be missing in another data set. There may also be temporal differences because the input data sets that were used for the reconstruction come from another date. Typically, there is no plan to maintain and update the once generated 3D data. This may be another source for indiscrepancies. All these differences have profound influences in practice, such as affecting the applications for which an existing 3D model can be used, the processing that is necessary to use it, and the likely errors that will be present in the end result. In this research we demonstrate how to create a consistent country wide 3D city model in LoD 1.2, 1.3 and 2.2. In order to achieve this, we look at three main aspects.

First, to ensure consistency between 3D city models of the same area and different LoD's, improve efficiency and serve the 3D data needs of different urban applications, we investigate how to reconstruct building models for large areas at different LoDs in one reconstruction process, based on the same reconstruction principles and based on the same source data. For the block models, we provide the user with several reference heights, so that the user can select the appropriate reference height to extrude building blocks for the specific application.

Second, our objective is to develop a fully automated reconstruction method. Our focus is on 3D city models covering large areas to support countrywide urban applications. This requires a fully automated reconstruction method. Automated reconstruction also enables standardisation of the output data resulting in consistent geometries, semantics and temporal aspects of the data. This consistency is also ensured when new models are reconstructed in the future with the same automated procedure based on updated source data.

Third, we investigate how to monitor and assess the quality of the building models that are automatically generated on such a large scale. This is essential for users to assess if models are fit for a specific use.

Finally, we also investigate the visualisation and dissemination of such a big dataset so that the city model is accessible and usable in an efficient manner.

**Structure of this paper**
In this paper we present our methodology to reconstruct LoD1.2, LoD1.3 and LoD2.2 models of all buildings in the Netherlands in one process. Section 2 further outlines the scope of this research by elaborating on the different LoDs of 3D city models and related work. We present our reconstruction methodology in Section 3 and the implementation in Section 4. We close with conclusions in Section 5.

## 2. SCOPE OF THE RESEARCH AND PREVIOUS WORK

In this section, we first explain how the reconstructed level of detail of 3D building models depends on the one hand on the data requirements for which the data is collected (Section 2.1) and on the other hand on the reconstruction method (Section 2.2). We then describe the motivation for the LoDs that we reconstruct in our research in Section 2.3. Finally, Section 2.4 presents other work on the reconstruction of 3D data for large areas.

### 2.1 LoD in relation to urban applications

As in 2D, a one-fits-all approach does not exist for a 3D city model. Instead specific applications require specific 3D data as is analysed in Biljecki et al (2015). For example, block models (LoD1) are sufficient for shadow-, wind- and noise- simulations. Roof structures (LoD2) with information on the roof materials are needed for solar potential estimation or in accurate energy demand estimation. Although LoD2 models are often also preferred in visualisations since they provide a realistic experience, realistic looking LoD2 models could still be ambiguous (Biljecki, 2018).

## 2.2 LoD in relation to reconstruction method

LoD1 models for every building can be automatically generated rather easily from point clouds and 2D building polygons, i.e., footprints (Ledoux et al 2021). Therefore, LoD1 models are frequently generated by various organisations as such source data are increasingly available as open data. However, automatically generated LoD1 models for the same area can still differ in for example their reference heights (e.g. the rooftop, the gutter height, one third of the roof-height) and the underlying statistical calculations. Many users are not aware of those multiple options to reconstruct a 3D block model, while these options do influence the outcome of analyses for which the LoD1 models are used (Biljecki et al., 2018). With respect to LoD2 models, many roof shapes can be generated fully automatically, although LoD2 reconstruction is still a current topic of research, as both the quality of available surveyed data and new 3D reconstruction algorithms still steadily improve (Rottensteiner et al. 2014; Lafarge et al. 2015; Pârvu et al 2018).The additional elements for LoD3 models are hard to reconstruct in an automated manner. Therefore, they are generated manually or the result of converted IFC models from the BIM domain (Donkers et al, 2016; Colucci et al 2020).

## 2.3 The LoDs in our research

In our research we focus on the reconstruction of LoD1.2, LoD1.3 and LoD2.2 building models using the terminology of the refined LoD framework of Biljecki et al (2016). We distinguish between two types of LoD1 models: LoD1 models that are a result of extruding a complete building footprint to one height, i.e. LoD1.2 models in Biljecki et al. (2016) and models that are extruded to one or more heights in case significant height jumps occur within the footprint, like a church with a tower or a house with a shed attached, i.e. LoD1.3 models. Both models are relatively simple and are therefore appropriate for applications that need simplified models. But LoD1.3 models enable more realistic visualisations. In addition, LoD1.3 models are also more accurate data for simulations that take block-shaped models of buildings as input, such as noise simulation where buildings act as noise barriers. This is why the automated reconstruction of LoD1.3 models is included in our research. LoD1.3 models are more difficult to automatically generate than LoD1.2 because it requires the detection of height discontinuities within the building footprint. The LoD1.0/LoD2.0 and LoD1.1/LoD2.1 models are based on generalised footprints and therefore outside the scope of our research. The LoD3 representations are outside our scope since they require manual work.

## 2.4 3D city models of large areas

There are many other examples of data sets containing building models of large cities or even nations as shown by an inventory by Santhanavanich (2020). Examples are the whole USA containing 125 million building models at LoD1 and the city of New York (with 100 LoD2 models of iconic buildings), as well as the LoD2 models of Montreal, Helsinki, Singapore, cities in North Rhine-Westphalia State (in LoD 1 and LoD2), and many other cities in Germany. An example of a LoD2 building data set covering a whole nation is the swissBUILDINGS3D 2.0 data set (swisstopo, 2021). It is a vector-based dataset which describes (among other topographic objects) buildings as 3D models with roof geometries and roof overhangs. The building models were extracted in a semi-automated manner from aerial images using a photogrammetric method of digital image (stereo) correlation, enhanced with additional information as attributes. Other building elements (footprint, facades, roof overhangs) are created with automated procedures.

Several of these initiatives highlight a problem of existing 3D models: often they are the result of a one-time capture, with a few mostly manual updating exceptions. Updates and extensions may be considered in the future but are not foreseen at the moment they were captured. In addition, existing models resulting from the same workflow (and thus consistent) are limited to one or the most to two different levels of detail for the same area and therefore the 3D data is limited to specific applications. More often, different LoDs of the same area are a result of different workflows and are therefore non-consistent with respect to geometry, temporal aspects and semantics. Finally, detailed metadata about how the buildings were reconstructed including quality information are often not generated and thus missing.

## 3. METHODOLOGY FOR AUTOMATED 3D RECONSTRUCTION

In this section we describe the reconstruction methodology of LoD1.2, LoD1.3 and LoD2.2 that we have developed and implemented for large areas, which we deliver as both 3D models and 2D+heights data, see Figure 1. First, we describe the input data that we use in the reconstruction. Then, we describe the process itself. Since the LoD2.2 reconstruction generates information that is used in the reconstructions of the other LoDs (e.g. distinguishing between points that fall on walls and on roofs; generating a planar partition of the original footprint based on the identified roof planes), we start with the LoD2.2 reconstruction.

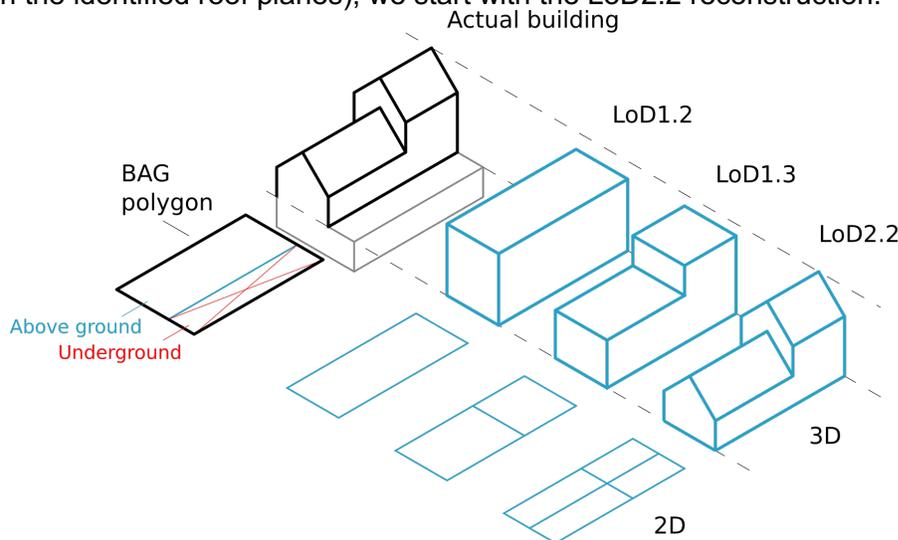

*Figure 1 Overview of the six representations that we reconstruct in our process. BAG stands for "Basisregistratie Adressen en Gebouwen", or the Building and Address register of the Netherlands; LoD is level of detail.*

### 3.1  Input data

For our reconstruction process we use building polygons (BAG) and LiDAR point data (AHN) as input.

**BAG:** The polygons come from the Building and Address register of The Netherlands (BAG). This data set contains all buildings and addresses in the Netherlands. The geometry of addresses is collected as points and those of buildings as polygons (i.e. outline as seen from above). Municipalities are responsible for collecting the BAG data and keeping the data up-to-date. The geometry for BAG buildings is also acquired from aerial photos and terrestrial measurements and the data positional data accuracy is 30cm. The data is provided via the national geo portal PDOK (2021) both in a viewer and as download service. BAG also contains the history of buildings, i.e. buildings that are planned and buildings that have existed in the past but now have been removed. For our reconstruction we make a selection of the BAG buildings that have been realised and have not (yet) been demolished, i.e. the input building data set represents the current situation.

**AHN:** The national height model of the Netherlands (AHN, 2019) is a point cloud acquired by airborne lidar. The first version of AHN (with a density of at least one point per 16 square meters, and in forests one point per 36 square meters) was completed in 2003. In the period of 2009 to 2012, the second version of the data set was acquired with an average point density of 10 points per square meter. The third version has been collected between 2014 and 2019. The resolution of AHN3, that we use for our reconstruction process, is similar to the one of AHN2. In addition, it contains a classification of the point cloud. For the AHN2 and AHN3 point clouds it is specified that an object of 2 × 2 m can be mapped with an accuracy of at least 50 cm. The height accuracy is 10 cm. We use the classes "building" and "ground points" to determine building heights respectively heights at ground level. The fourth version of AHN will become available in the next two years. AHN4 will have a point density of about 10-14 points per square meter, and in some locations even higher.

### 3.2    LoD2.2

Our reconstruction method improves upon our earlier research as described in Dukai (2019, 2020) and Stoter et al (2020). The main improvement in this work is the addition of LoD2.2 output.
Our method uses footprints and height points that are well aligned as input and consists of two steps. In the first part the input footprint is partitioned into roof parts. And in the second part this 2D roof partition is extruded into a 3D model.

**Footprint partitioning**
In this first step, the input footprint is partitioned by breaklines detected in roof planes (see Figure 2).

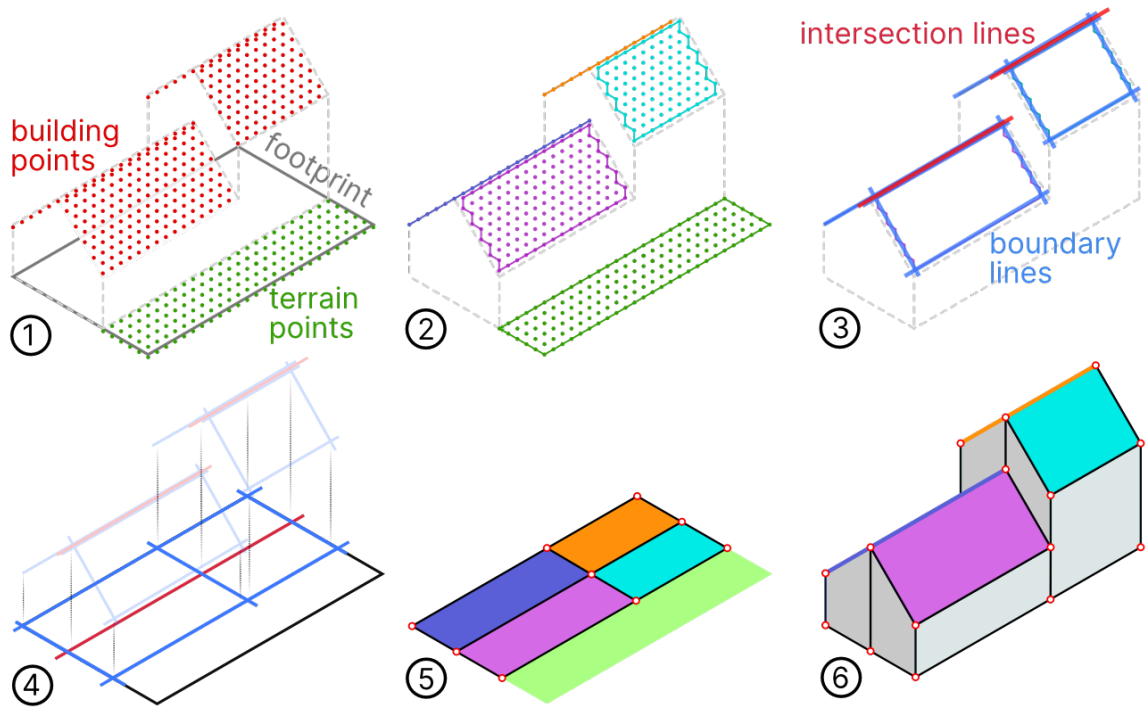

*Figure 2 The main steps in the reconstruction process: 1) Building-polygon + AHN- surface points. 2) roof plane detection. 3) line detection. 4) lines are projected into initial partition. 5) final partition after assigning roof planes to polygon (compare with 2). 6) LoD2.2 3D mesh*

These roof planes are detected if sufficient points can be found for that plane using a region-growing algorithm (see Figure 2.2). For the AHN that we use, with a point density of ~8 points/m2, we set the minimum number of points to 15 which is equal to a roof element of about 2 square meters. Points that are on a wall plane (facade) or not part of any plane are removed. We derive two types of lines from the planes: boundary lines and intersection lines (see Figure 2.3). The boundary lines of roof planes are detected using the α-shape of each detected roof plane. The intersection lines are generated at the location where adjacent planes intersect, e.g. on top of a gable roof.

Before the boundary and intersection lines are used to subdivide the footprint, they are regularised and duplicate lines are removed. For example, the line on top of the gable roof in Figure 2.3 is detected three times: once as an intersection line and twice as a boundary line, i.e. once for each incident roof plane.

The remaining lines are used to subdivide the footprint into an initial planar partition (Figure 2.4). This is referred to as the *initial roof partition*. The *initial roof partition* may still have a high complexity, i.e. it may contain many small faces. To further reduce the complexity of the roof-partition, an optimisation step is performed using a graph-cut optimization (Zebedin, 2008). In this step a roof plane is assigned to each face in the roof-partition (see Figure 2.5). This is done in such a way that the total error with the input point cloud is minimised and the total length of the edges between faces of a different roof plane is minimised. The latter assures a minimum number of edges and vertices, i.e. a low model complexity. After this step, the edges for which the two incident faces are assigned to the same roof plane are removed from the partition. The faces in the resulting *final roof partition* are referred to as *roof parts*.

**Extrusion**

In the LoD2.2 reconstruction the identified roof parts are extruded from ground level to a 3D mesh (Figure 2.6). The mesh consists of three types of surfaces: the ground plane, the roof surfaces and the wall surfaces. The height of the ground plane is based on the lowest point around the building and is calculated as the 5th percentile of all ground points that are within a 4m buffer of the building. An intersection curve of the terrain could also be used for this. The construction is done in such a way that no internal walls are created and the mesh is topologically correct.

### 3.3     LoD1.3

The LoD1.3 reconstruction uses the same footprint partitioning as is generated for the LoD2.2 reconstruction. But for LoD1.3, the footprint partitions are further simplified by merging neighboring parts that have no significant height difference. We use 3m as a threshold in this process, which is more or less equal to a floor-height, no matter the area. As the height reference we use the 70th percentile height for each roofpart. The merging starts with merging from small to large height gaps and is an iterative process, i.e. if merging two roofparts leads to an elevation difference of <3m with another part they are merged again. The iteration stops when there are no more height differences smaller than 3m between adjacent roof parts.
In a next step, reference heights are calculated for each remaining part and used to extrude the part.
As explained in Section 2.2, these reference heights can represent different extrusion heights for one building, i.e. the roof edge, the ridge height or the maximum height (like a chimney). Furthermore, the underlying statistical calculations used to calculate the extrusion height can differ, e.g. average, median or maximum.
To standardise possible extrusion variations and to let the user choose which one to use, our method calculates four different reference heights from the points that fall on a roof part (excluding the points on walls) and assigns these to the 2D roof-parts, i.e. minimum, maximum, 50th percentile and 70th percentile.
The LoD1.3 models are provided in two representations: as 2D roof-parts with the different reference heights as attributes and as reconstructed (i.e. extruded) 3D models based on the 70th percentile of the roof points the specific part contains. The 2D roof parts also contain the ID of the original building to link the individual roof parts to the original building. The 3D models are reconstructed in such a way that there are no inner walls.
The reconstructed 3D models are represented as solids since rules for solids are stricter than for MultiSurfaces e.g. using the solid geometry enforces a 2-manifold (i.e. watertight) 3D object. For the ground surface, the same height is used as for the LoD2.2 models, i.e. 5th percentile of all surface points that fall within 4m radius of footprint.

### 3.4     LoD1.2

For the LoD1.2 models we calculate the same four reference heights for the extrusion as we do for LoD1.3 and assign these to the original footprints. Also for LoD1.2, our method both reconstructs the 3D block models (based on the $70^{th}$ percentile of height points that fall on all roof parts of the building) and assigns the four reference heights to the original 2D footprints. The same surface height is used as for the other LoDs.

## 4. IMPLEMENTATION

### 4.1 Implementation of reconstruction process

The implementation of the whole process as described above is visualised in Figure 3. The input data is tiled to make the reconstruction and dissemination of the data more manageable. After reconstruction the building models are stored in a PostgreSQL database from which the data is exported or directly consumed in various formats.

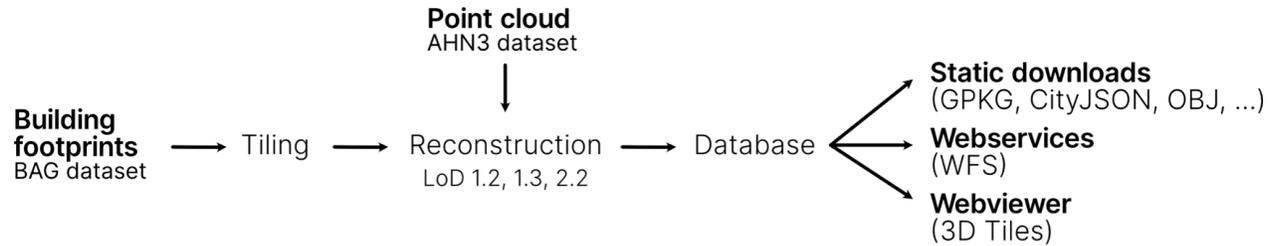

*Figure 3 Overview of the multiple LoD reconstruction process*

There are two considerations for optimally tiling the building footprints. First, the objective is to limit the number of buildings in each tile, so that the workload is as balanced as possible between the processes. Second, the buildings should be spatially clustered, so that the corresponding point cloud can be read efficiently. To meet both conditions, we use a quadtree with a maximum cell size of 3500 for subdividing the buildings (see Figure 4).

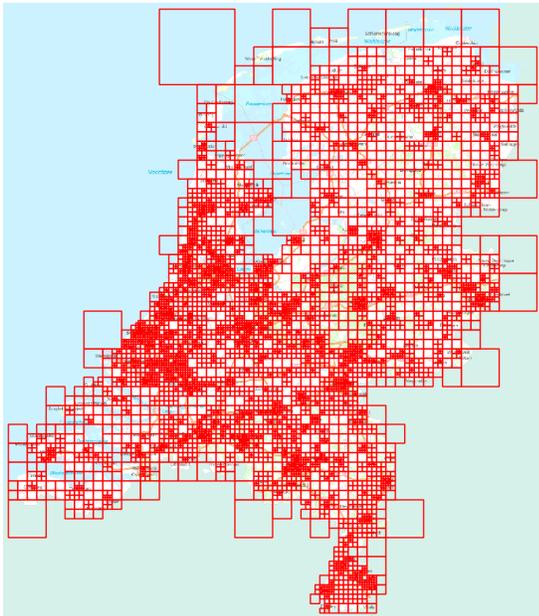

*Figure 4 Quadtree-based tiling scheme for data processing and dissemination*

Thus, the building tiles are the leaves of the quadtree, where each tile contains a maximum of 3500 buildings. This assures that the reconstruction-time per tile is more or less the same and that the tiles available for download are similar in file size. The reconstruction of all ten million buildings in The Netherlands takes about 40 hours, with 30 concurrent processes on a single machine (2 x Xeon E5-2650 CPU, 128GB RAM). The computation cost scales linearly with the number of buildings, since each building is processed independently. The reconstruction process is highly automated, which allows us to quickly run a new iteration in case of improvements in an algorithm or in case new input data becomes available. Figure 5 shows an example of reconstructed 3D models at different LoDs for one building.

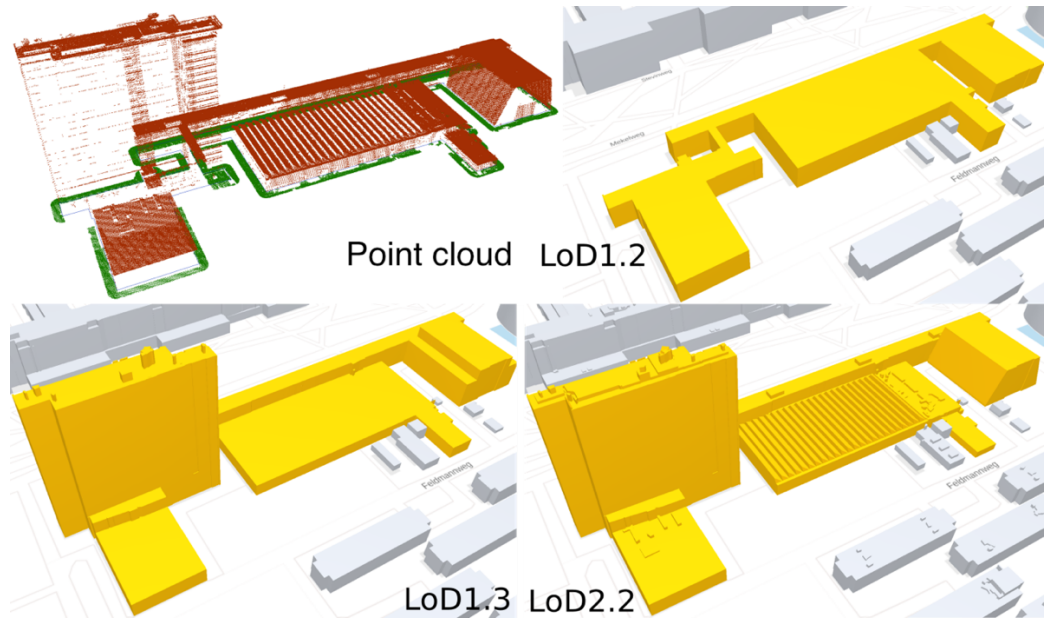

*Figure 5 Faculty building of Electrical Engineering, Mathematics and Computer Science at TU Delft campus. AHN3 point cloud and reconstructions at LoD1.2, LoD1.3 and LoD2.2. At LoD1.3 only height jumps >3m are kept and therefore it contains less roof structures than LoD2.2*

**4.2 Visualisation and dissemination**

To view and query as well as to download the reconstructed building models, we built a website with a 3D viewer (Figure 6). The 3D viewer was developed with two main goals: network performance (ie. fast fetching of the data) and client performance, i.e. to minimise the resource needed on the device being used (including mobile devices). We developed our own solution, since we could not find a suitable off-the-shelf solution.

To satisfy the network and client performance, we use a web graphics friendly format with minimal size requirements, i.e. 3D Tiles[1]. This is based on the glTF[2] format. We export the dataset to 3D Tiles using the same tiling scheme as for the reconstruction of the buildings, which ensures that tiles have a relatively equal distribution of objects. We use 3DTilesRendererJS[3] to render these tiles in the viewer. Our solution to make the user interface easy for users and mobile friendly uses VueJS[4] for the website's logic and Bulma[5] for the styling of the user interface elements. The viewer provides several functionalities that enable users to investigate the whole dataset, as well as share it with others.
A user lands at an initial point and can move around the country.

---

[1] https://github.com/CesiumGS/3d-tiles

[2] https://www.khronos.org/gltf/

[3] https://github.com/NASA-AMMOS/3DTilesRendererJS/

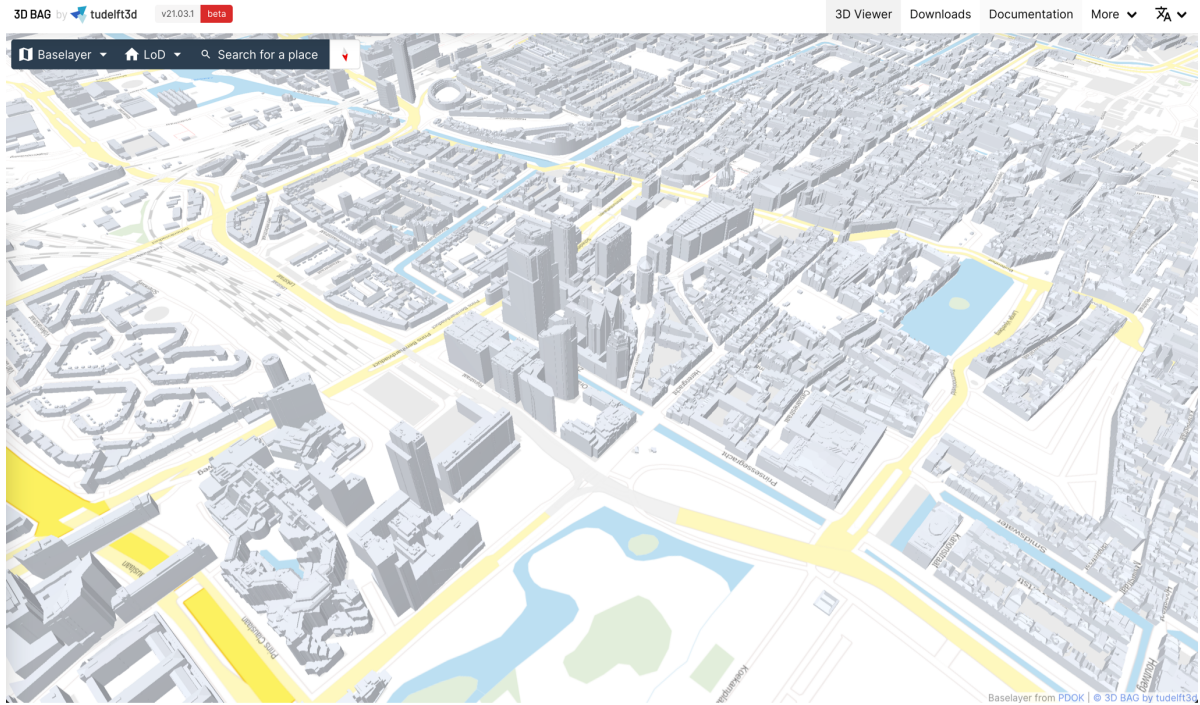

*Figure 6 Screenshot of the developed 3D viewer available at www.3dbag.nl.*

A simple flat terrain using WMTS[6] is used to provide proper orientation context to the user. Each location in the viewer corresponds to a unique address, so that the user can bookmark or share the current view with others. Finally, the user can click on a building and get its information, as well as some derived properties: the height of the building at the specified point, and the slope of the surface that the user clicked on.
Through the website 3DBAG.nl, the data is downloadable in different formats: GeoPackage, PostGIS backup, Wavefront OBJ and CityJSON. The viewer also contains a function to report errors by users to help us improve our process which can concern any part of the process, i.e. from preprocessing input data to reconstruction, viewing and use.

### 4.3 Quality information

Quality information regarding the resulting models is needed to identify a badly reconstructed model or an exceptional situation for which the 3D reconstruction process had not yet accounted for. With this information the reconstruction process can be improved. In addition, it provides the user with information on how good a specific model is so that the user can act upon this.

---

[4] https://vuejs.org/

[5] https://bulma.io/

[6] http://opengeospatial.github.io/e-learning/wmts/text/index.html

We calculate two types of quality parameters and assign these as attributes to the individual models. First, we calculate parameters that assess the quality of the source data for the specific building. For example, the number of points that were available for the 3D reconstruction, the no-data area, and the timeliness of the source data. Second, we calculate parameters that measure the success of the automatic reconstruction, e.g. the Root Mean Square Error between the reconstructed model and the input points, the maximum error between reconstructed mesh and the point clouds and eventual invalidity codes both in 2D (which means the input data contained an error) and in 3D (which means that the reconstruction failed). For the LoD2.2 building models of the Netherlands, the RMSE is less than 31 cm for 95% of the models and less than 9 cm for 75% of the models (see Figure 7). A more extensive evaluation is in progress. This evaluation is based on the quality parameters that we calculate in order to identify opportunities to further improve our workflow (Dukai et al, 2021).

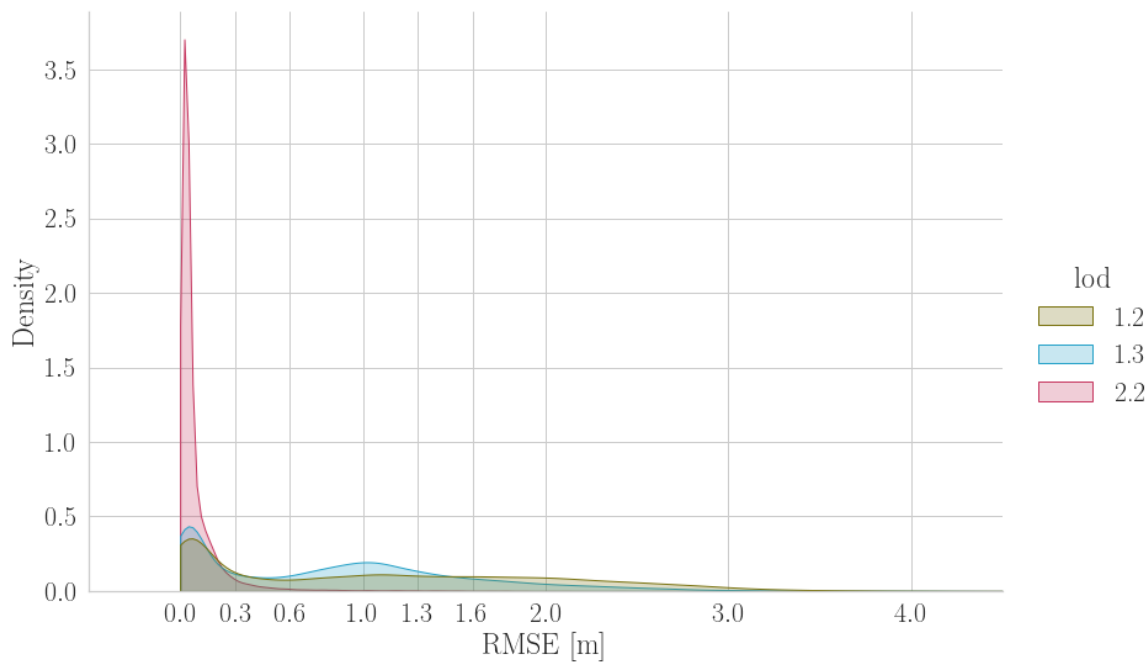

*Figure 7 The root mean square error between the input point cloud and the LoD2 reconstruction result (Dukai at al, 2021).*

## 5. CONCLUSIONS

In this paper, we describe the process that we have developed to automatically reconstruct LoD1.2, LoD1.3 and LoD2.2 building models (supporting different reference heights for the block models) for large areas in one reconstruction process and based on the same source data. This provides the user with consistent 3D data of the same area meeting the data-needs of different applications. We monitor quality information throughout the whole process in order to continuously improve the process from input data, pre-processing and 3D reconstruction to download and use the data in urban applications. In addition, the user can use the quality information to decide on the fit-for-purpose of the data for their own application.

The 3D data that we generate has been a good source to experiment and test all kinds of urban applications that need 3D data and it is being used in for example noise simulations (Stoter et al, 2020), wind flow simulations (García-Sánchez et al, 2021) and energy consumption calculations (Wang et al, 2020; León-Sánchez et al, 2021).

Based on experiences and users' feedback we will improve the process and optimize for different users and applications. These improvements may be generic such as filling the no_data areas in the point cloud with artificial intelligence; better alignment to specific data needs of urban applications like optimising the level of detail for specific applications (in relation to processing time); and, enriching the data with relevant information for example for distinguish between external and internal walls and their areas for energy-related applications or the estimation of number of floors per building. In addition, the availability of the next version of AHN (to be expected next year) will provide more reconstruction possibilities as can be seen in Figure 8. In following research, we will also study more fundamental issues i.e. how to maintain and manage different temporal and geometric versions of the 3D data and how to better align the different parts in the process to obtain even better results.

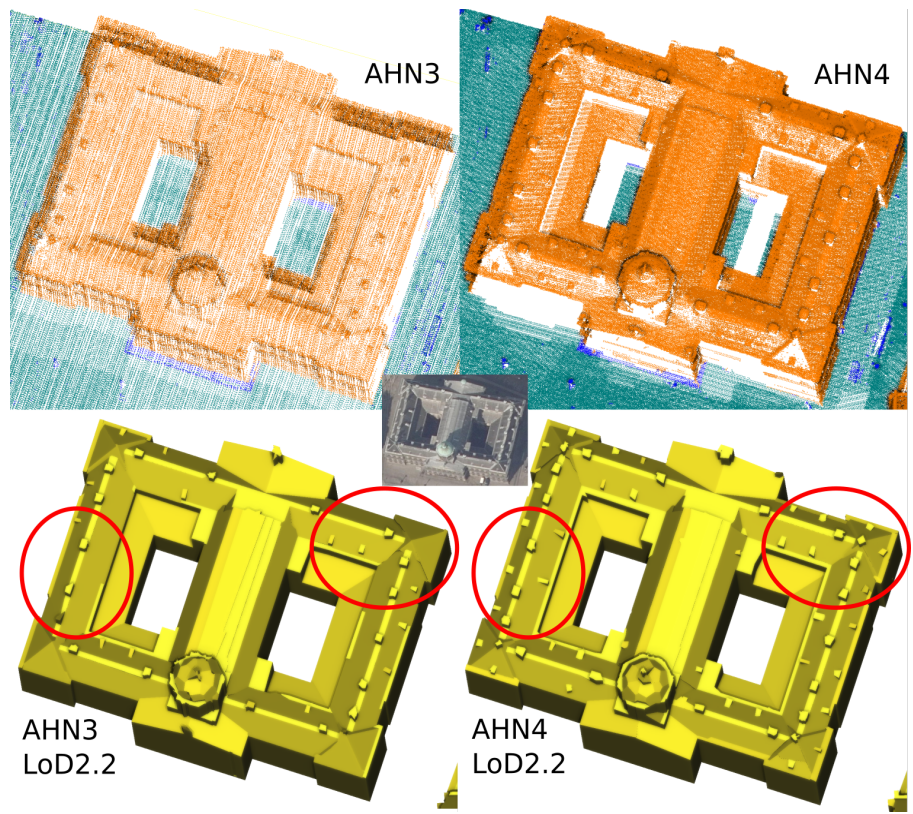

*Figure 8 The Palace in Amsterdam. AHN3 and AHN4 point clouds and the resulting LoD2.2 reconstruction. More details are reconstructed from the AHN4 point cloud because it has a higher resolution.*

**ACKNOWLEDGEMENTS**
This project received funding from the European Research Council (ERC) under the European Union's Horizon2020 Research & Innovation Programme (grant agreement no. 677312 UMnD: Urban modelling in higher dimensions). Part of the research was done within the VOLTA project funded from the European Union's Horizon 2020 research and innovation programme under the Marie Skłodowska-Curie (grant agreement No 734687).